\documentclass[11pt]{article}

\usepackage[final]{acl}
\usepackage{tabularx}
\usepackage{times}
\usepackage{latexsym}
\usepackage{ragged2e}
\usepackage{makecell}
\usepackage[T1]{fontenc}

\usepackage[utf8]{inputenc}

\usepackage{microtype}

\usepackage{inconsolata}

\usepackage{graphicx}

\title{Evaluating FrameNet-Based Semantic Modeling for Gender-Based Violence Detection in Clinical Records}

\author{
Lívia Dutra\textsuperscript{1,2}, 
Arthur Lorenzi\textsuperscript{1,3},
Frederico Belcavello\textsuperscript{1},
Ely Matos\textsuperscript{1}, 
Marcelo Viridiano\textsuperscript{1}, \\
\textbf{\large{Lorena Larré\textsuperscript{1},
Olívia Guaranha\textsuperscript{3},
Erik Santos\textsuperscript{3},
Sofia Reinach\textsuperscript{3}, Pedro de Paula\textsuperscript{3},
Tiago Torrent\textsuperscript{1,4}}} \\
\textsuperscript{1}Federal University of Juiz de Fora (FrameNet Brasil), 
\textsuperscript{2}University of Gothenburg \\
\textsuperscript{3}Vital Strategies Brasil, 
\textsuperscript{4}Brazilian National Council for Scientific and Technological Development (CNPq) \\
\texttt{livia.vicente.dutra@svenska.gu.se}, \texttt{\{marcelo.viridiano, lorena.tasca\}@estudante.ufjf.br}, \\ 
\texttt{\{alorenzi, oguaranha, esantos, pcbpaula, sreinach\}@vitalstrategies.org}, \\
\texttt{\{fred.belcavello, ely.matos, tiago.torrent\}@ufjf.br}
}

\begin{document}
\maketitle
\begin{abstract}

Gender-based violence (GBV) is a major public health issue, with the World Health Organization estimating that one in three women experiences physical or sexual violence by an intimate partner during her lifetime. In Brazil, although healthcare professionals are legally required to report such cases, underreporting remains significant due to difficulties in identifying abuse and limited integration between public information systems. This study investigates whether FrameNet-based semantic annotation of open-text fields in electronic medical records can support the identification of patterns of GBV. We compare the performance of an SVM classifier for GBV cases trained on (1) frame-annotated text, (2) annotated text combined with parameterized data, and (3) parameterized data alone. Quantitative and qualitative analyses show that models incorporating semantic annotation outperform categorical models, achieving over 0.3 improvement in F1 score and demonstrating that domain-specific semantic representations provide meaningful signals beyond structured demographic data. The findings support the hypothesis that semantic analysis of clinical narratives can enhance early identification strategies and support more informed public health interventions.

\end{abstract}

\section{Introduction}

The World Health Organization estimates that one in three women experiences physical or sexual violence by an intimate partner at some point in her life \cite{WHO2024}. Gender-based violence (GBV) is therefore not only a social issue, but a major public health concern  \cite{garcia2011violence,sweet2014every,ohman2020public}. In Brazil, healthcare professionals are legally required to report cases of violence. Yet underreporting remains widespread. Either because victims are unable or unwilling to report their experiences or because signs of violence go unrecognized within routine medical encounters. Research suggests that many professionals struggle to identify signs of abuse, lack appropriate support tools, and work within fragmented information systems that do not communicate effectively  \cite{kind2013subnotificacao, saliba2015desafios}.

Brazilian public health systems collect large amounts of data on hospitalizations, mortality, medical records, and violence notifications. However,  these systems are not fully integrated and lack a shared individual identifier \citep{guaranha2025data}. As a result, it is difficult to follow trajectories of risk over time or across institutions. Most of this information is stored in parameterized fields, which facilitate statistical analysis but capture only structured aspects of clinical encounters. Electronic medical records, however, also include open-text fields where healthcare professionals describe symptoms, circumstances, and patient histories in more detail. These narrative records often contain rich descriptions of situations that may signal risk of violence, but they are rarely consider for analysis due to its complexity. 

The research reported in this paper explores whether semantic analysis of clinical narratives can help identify potential cases of GBV earlier and more reliably. The underlying assumption is that linguistic patterns embedded in medical records may reveal indicators of violence that are not captured by structured data alone. In particular, we investigate the contribution of FrameNet-based annotation to identifying possible patterns of violence within routine health data.\footnote{This study is based on the first authors' master's thesis and has been presented as part of a book chapter to lustrate the social applicability of FrameNet \cite{BPLN_livro_cap-framenetbr}.} To achieve that, three experimental setups are compared using a SVM classifier : (1) a model trained on manually and automatically frame-annotated open-text data; (2) a model trained on annotated open-text combined with parameterized data; and (3) a model trained exclusively on parameterized information. Model performance is assessed using precision, recall, and F1-score, along with qualitative analysis of possible semantic patterns.

The results show that models incorporating FrameNet-based semantic information outperform those relying solely on structured data, achieving an F1-score of 0.772, compared to 0.461 for the model trained exclusively on parameterized data. This result, backed by the qualitative analysis of the findings, suggests that semantic analysis of clinical narratives can provide meaningful support for the identification of gender-based violence in primary healthcare settings.

\section{Frame-Based Models of Linguistic Cognition}

FrameNet is a corpus-based computational lexical database grounded in Frame Semantics, a theory within Cognitive Linguistics proposed by  \citet{Fillmore1982}. In Frame Semantics, word meaning is not treated as self-contained. Instead, meaning is understood through mental representations, called \textit{frames}, which capture shared knowledge about recurrent situations, the participants involved in them, and the relations between those participants. A \textit{frame} can therefore be seen as a structured background against which individual words are interpreted. Understanding a lexical item presupposes familiarity with this wider conceptual structure, since the meaning of any single element depends on how it fits into the scenario as a whole.

\begin{figure}
    \centering
    \includegraphics[width=1\linewidth]{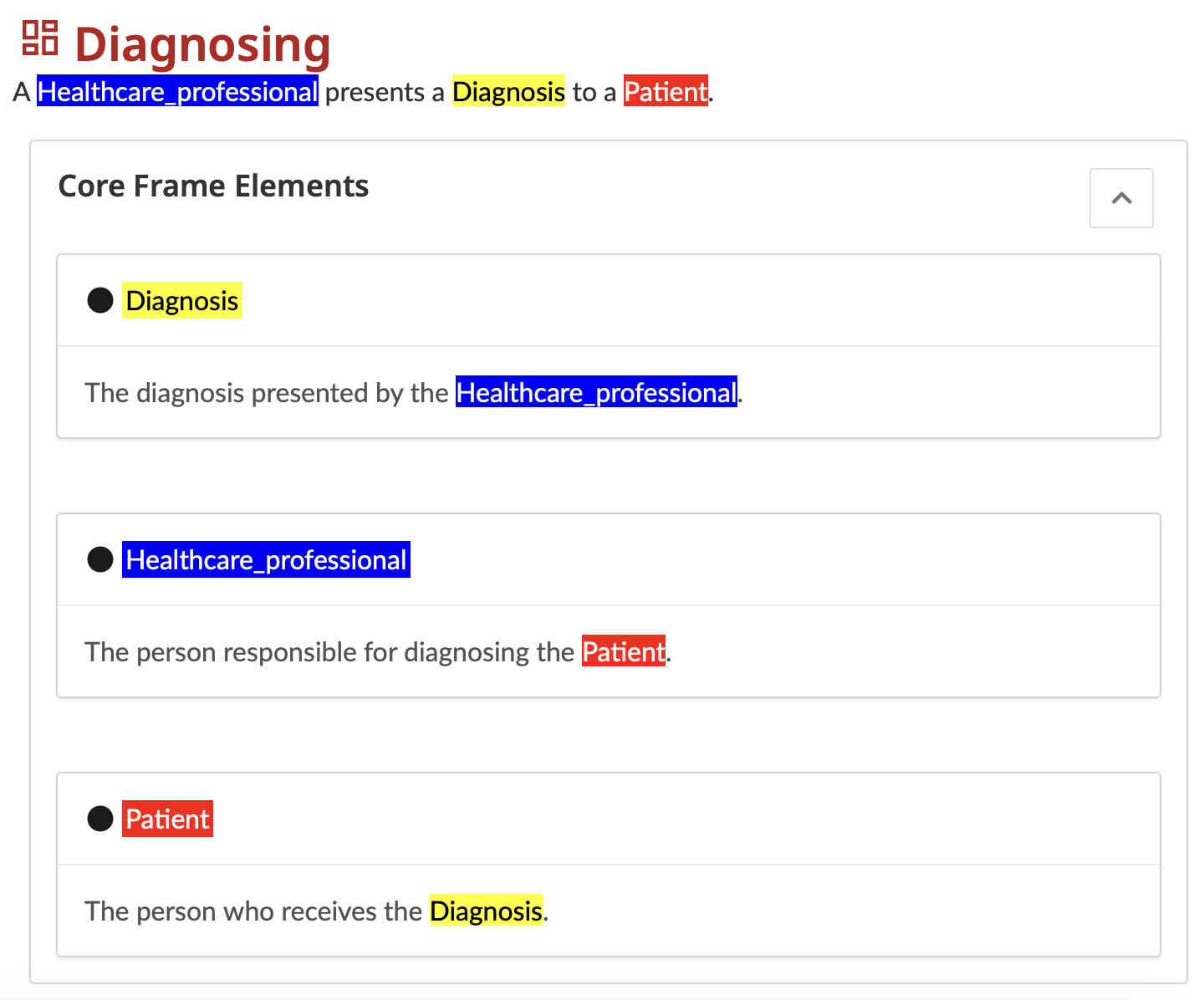}
    \caption{The \texttt{Diagnosing} frame.}
    \label{fig:Frame}
\end{figure}

For instance, consider the lexical unit \textit{diagnose.v}, which evokes the \texttt{Diagnosing} frame –– Figure \ref{fig:Frame}. The verb does not simply refer to an action. Rather, it presupposes a healthcare context in which several roles are necessarily present. At a minimum, there must be a healthcare professional who makes the diagnosis and a patient whose condition is being evaluated, defined as Frame Elements in the theory. Without these participants, the situation would be difficult to interpret as a diagnosis. The frame also allows for additional elements, such as the method in which the diagnosis was performed or the time and place that it happened. Thus, when the lexical unit \textit{diagnose.v} appears in a clinical record, it activates a rich scenario of healthcare assessment. 

FrameNet was then created in 1997 to implement this theoretical framework in a systematic, corpus-based resource for English. The model has since been extended to several languages, including Brazilian Portuguese through FrameNet Brasil (FN-Br) \cite{torrent-et-al-2022}. Using its annotation tool \cite{torrent2024flexible}, FN-Br is able to link lexical units to the frames they evoke and annotate the semantic roles associated with those frames using authentic language data. In doing so, it is possible to gather insights and capture semantic representations that are both cognitively motivated and computationally interpretable for broader and more specific domains, as is the case of the current study, which focuses on semantic representations of healthcare and violence narratives. By modeling these domains explicitly and annotating relevant corpora, it becomes possible to identify recurring semantic configurations and relational patterns that may not be immediately visible in surface-level, as it is discussed next.

\section{Corpora and Methods}

This section presents the corpora and methods used in this study. The methodology involved modeling specific domains, manual and automatic annotation of data, and the design of GBV identification models and their quantitative and qualitative evaluation, as described next. 

\subsection{Corpora}

The corpora used in this study were collected from health records produced in Recife, the capital of Pernambuco, Brazil, in collaboration with the Municipal Health Department. Data come primarily from two public health information systems: SINAN (Notifiable Diseases Information System) and e-SUS AB (Primary Healthcare e-Medical Records System). In addition, definitions from the International Classification of Diseases (ICD) were incorporated to interpret diagnostic codes appearing in medical records. Also, causes of death were extracted from the Mortality Information System (SIM) as an indicator of true positive cases of GBV, in cases of violent deaths. The dataset is extensive, comprising more than three million records in e-SUS and more than 13,000 records in SINAN.

Regarding the systems structure, SINAN contains both parameterized fields and one open-text field (“observation”) describing episodes of violence. For this study, only two components were used: (i) the parameterized indicator of a positive violence case and (ii) the corresponding open-text description. The e-SUS AB system records primary healthcare data through a combination of parameterized fields (e.g., age group, race, gender identity) and open-text fields following the SOAP model (Subjective, Objective, Assessment, Plan), along with additional narrative fields such as reason for referral, complements, and observations. Because diagnostic codes often appear without textual definitions in open-text fields, ICD descriptions were linked to the e-SUS records to enrich their semantic interpretability.

Given the presence of highly sensitive personal information in the corpora, strict ethical and legal safeguards were implemented as well as the anonymization of the data. The anonymization process combined automatic (NER models \cite{souza2019portuguese, pierreguillou2021nerbert,make4010003}, fuzzy search of local place names, and regular expressions), semi-automatic (frequency-based detection of potential names), and manual verification methods to ensure the removal of Personally Identifiable Information (PII).  Once this process was completed, anonymized samples of the data were used to model specific domains and for frame-based annotation. 

\subsection{Domain-Specific Modeling } 

As a means to fully capture the narratives in the open-fields of the public information systems, the specific semantic domains of Healthcare and Violence were modeled in FN-Br. This process entails the structuring of a cognitive representation that connects essential concepts of a given topic. This process consists of a twelve-step methodology, involving not only corpora collection and anonymization, but also lexical expansion, the modeling of new frames and relations between frames and lexical units -- named Ternary Qualia Relations\cite{torrent2024flexible} –– and the lexicographic annotation of the corpora \cite{ruppenhofer2016framenet}, as described in \citet{dutra-et-al-2023} and \citet{larre-torrent-2024}. This process was carried out by a group of ten researchers and resulted in 35 frames and 2,776 lexical units for the Healthcare domain along with 48 frames and 1,774 lexical units for the Violence domain.

\subsection{Annotation}

The annotation process was carried out in two phases: human and automatic. Human annotation followed the FrameNet methodology \cite{ruppenhofer2016framenet} and was conducted using a sample of anonymized corpora with two aims. First, as the final stage of domain-specific modeling, annotation served the purpose of validating the domains; second, it was used to compose a dataset to train an automatic semantic labeler to be used in the entirety of the corpora. A total of seven trained annotators were part of the human annotation effort, which was carried out in a mirrored version of the FN-Br annotation tool \citep{torrent2024flexible} with gated access to the data -- so as to add one more layer of protection to the data being handled. The process focused on semantic annotation and consisted of selecting the frame evoked by each lexical unit in the sample sentences, which, then, generated an Annotation Set that allowed for the tagging of the frame elements represented in that narrative. The final number of annotated sentences was 2,352, resulting in over 14,600 Annotation Sets. 

Automatic semantic labeling was performed using a newly trained version of LOME \cite{xia2021lome}, a multilingual information extraction system that integrates a FrameNet parser within a pipeline based on XLM-RoBERTa, a BIO tagger and a Typer. For this study, LOME was trained in FrameNet Brasil’s annotated data for English and Portuguese, including data from the Violence and Healthcare domains, expanding the training data used by \citet{xia2021lome}. The newly trained model had a micro-F1 of 50.68, slightly less than the original implementation (F1 = 56.34). Because this new instance was trained to deal with more challenging data, given its specificity, the performance is satisfactory. After training, the model was used to automatically annotate frames and frame frame elements in open-text sentences from e-SUS AB, SINAN, and ICD records.

\subsection{GBV Identification Models}

As stated previously, this study focuses on evaluating the use of FrameNet-based semantic annotation to identify GBV cases and patterns in open-text medical records. In this sense,  a model was developed to integrate FrameNet-annotated data and enable an assessment of feature importance for distinguishing violence from non-violence in e-SUS records.  

Thus, as a means of accomplishing that, three experimental setups were conducted.  All experiments used a linear Support Vector Machine (SVM). This choice was motivated by its interpretability, suitability for high-dimensional data, and consistency with the original project design. The three experimental setups used the same subset of e-SUS records, originally categorized based on ICD codes and links to SINAN notifications and SIM records. There were four labels:

\begin{itemize}
    \item Violence: records with an ICD code for aggression or within two days of a SINAN notification or SIM record with the same code;
    \item Non-violence: ICD codes that have a small probability of being associated with violence, e.g. COVID-19 and some congenital malformations;
    \item Likely Violence: any record within 30 days of a notification of violence that does not have an ICD code for aggression; 
    \item Unknown: any record that does not fall into one of the previous categories. 
\end{itemize}

For this study, only violence and non-violence records were used, resulting in 801 cases (634 non-violence; 167 violence). Non-violence cases were undersampled to reduce class imbalance. Additionally, two specialists reviewed 100 "likely violence" cases, who reclassified them as 17 violence and 83 non-violence, increasing the complexity of the dataset by including more ambiguous cases.

The three experimental setups designed were:

\begin{enumerate}
    \item \textbf{Semantic Model:} In the first setup, only the LOME annotated open-text fields were considered. As LOME identifies frame targets but not lexical units (LUs), an additional procedure to extract the LUs associated with each annotated span was also applied. This step increased the granularity of the representation, allowing the model to capture not only frames and frame elements, but also specific lexical choices. From these annotations, feature vectors were constructed based on the frequency of frames, frame elements, and lexical units, as well as the co-occurrence of frame elements across frames. Co-occurrences were considered only when at least one of the frames belonged to the Healthcare or Violence domains. The Ternary Qualia Relations between lexical units were also incorporated, assigning them a small weight to enrich the semantic connections without overwhelming the representation. To reduce sparsity, the least frequent features were removed (frames with fewer than 50 occurrences and LUs with fewer than 25). The resulting vectors were weighted using TF-IDF and L1-normalized. Given the high dimensionality of the semantic representation (15,456 features), Principal Component Analysis (PCA), was applied to reduce it to 2,000 components while preserving 94.8\% of the variance. These components served as input to the classifier. 
    \item \textbf{Mixed Model:} The second setup considered annotated open-text fields and selected annotated parameterized fields. This experiment followed the same pipeline as the first, but additionally incorporated structured parameterized fields into the semantic representation. Categorical variables –– such as race, gender identity, sexual orientation, prosthesis need, and age group –– were mapped to corresponding lexical units and frames. After TF-IDF weighting, the feature space comprised 15,478 dimensions. PCA again reduced this to 2,000 components, preserving 93.5\% of the variance. This combined representation was used as the input to the model.
    \item \textbf{Demographic Model:} The third setup excluded semantic annotation and relied exclusively on parameterized data. The features included demographic, clinical, and administrative variables such as race, age, ICD codes, marital status, education level, unit location, and referral timing. Categorical variables were transformed using One-Hot Encoding, expanding the original 20 structured variables to 142 binary features. These features were used directly as input to the classifier.
\end{enumerate}

\subsection{Evaluation}

After training, the model was evaluated both quantitatively and qualitatively. While the quantitative evaluation provided numerical evidence of model performance, the qualitative analysis aimed to interpret and understand what the models were learning and how FrameNet annotation contributed to GBV identification.

\paragraph{\textbf{Quantitative (SVM) Evaluation}}

To compare the three experimental setups and assess the relevance of different data sources for GBV identification, the performance of the models was evaluated using five-fold cross-validation. In this procedure, the dataset was divided into five subsets: in each iteration, four subsets were used for training and one for testing, rotating until all subsets had served as the test set. Performance was assessed using precision, recall and F1 score, and the final results correspond to the average across the five folds. 

\paragraph{\textbf{Qualitative Evaluation}}

This process involved two complementary lines of analysis: first, feature importance scores were extracted from the best-performing semantic model and the demographic model; second, the most frequently evoked frames and lexical units in the annotated e-SUS records of confirmed victims were also examined.

\begin{enumerate}
    \item \textbf{Model features:} The 35 most relevant features for both the semantic and the demographic models were analyzed based on their contribution to the classification of the cases\footnote{At this point, it is not possible to identify for which of the classes each feature was most relevant, only that they were relevant for the model's decision.}. This allowed us to assess the explanatory power of parameterized fields versus semantic features and to identify key frames, frame elements, and lexical units associated with GBV cases.

    \item \textbf{Frame and lexical pattern analysis:} Next, the frame activation patterns were analyzed in confirmed GBV cases by examining:
    \begin{itemize}
        \item the 15 most frequently evoked frames in both domains –– Healthcare and Violence;
        \item the 20 most frequent LUs per domain;
        \item the 30 most frequent LUs evoking the \texttt{Health\_conditions} frame, to explore possible links between health conditions and violence.
    \end{itemize}
    This analysis allowed  a better understanding of patterns that could be linked to GBV and pointed towards future investigation.
\end{enumerate}

\section{Results and Discussion}

In this section, we present the results of the evaluation conducted on the model setups and discuss their implications to the use of frame-based representations to the identification of GBV in e-medical records.

\subsection{Quantitative Evaluation: SVM Models} 

As shown in Table \ref{tab:model-performance}, the semantic model that relies solely on open-text fields consistently produced the strongest results. It obtained the highest recall, indicating an effective identification of positive cases. Precision was lower, but this trade-off is acceptable in the context of the study, once the main concern is avoiding false negative cases of violence in a setting where underreporting is an issue. The F1-score reflects this balance. Furthermore, recall values showed little variation across cross-validation splits, while precision and F1 varied more substantially, suggesting that some data splits were more challenging for the model than others.

Adding parameterized data to the semantic model did not lead to a meaningful improvement in the second experiment. Although a small increase in precision was observed, this was achieved without gains in overall performance and is not particularly advantageous for the task at hand, as it increases the likelihood of false negatives. These results reinforce the idea that semantic information extracted from textual descriptions is more important in identifying cases of violence than structured demographic attributes.

Finally, the contrast with the demographic model is clear. Although recall values were relatively high, precision was extremely low, indicating that many cases were incorrectly classified as positive. This imbalance makes the recall results difficult to interpret with confidence. Moreover, recall varied widely across different splits, resulting in consistently low F1-scores. Taken together, these results show that parameterized data alone are not sufficient for reliable case identification, even if they may be useful for descriptive analyzes.

\begin{table}[ht]
\centering
\renewcommand{\arraystretch}{1.3}
\setlength{\tabcolsep}{6pt}

\begin{tabularx}{\linewidth}{>{\RaggedRight\arraybackslash}p{0.25\linewidth}|c|c|c}
\textbf{Model} & \textbf{F1} & \textbf{Recall} & \textbf{Precision} \\
\hline
Semantic
& \makecell{0.772\\(0.113)}
& \makecell{0.838\\(0.071)}
& \makecell{0.756\\(0.190)} \\
\hline
Mixed
& \makecell{0.771\\(0.114)}
& \makecell{0.832\\(0.078)}
& \makecell{0.759\\(0.189)} \\
\hline
Demographic
& \makecell{0.461\\(0.089)}
& \makecell{0.701\\(0.173)}
& \makecell{0.345\\(0.057)} \\
\hline
\end{tabularx}

\caption{Model performance comparison}
\label{tab:model-performance}
\end{table}

Overall, the results support the initial hypothesis that FrameNet-based semantic analysis contributes meaningfully to the identification of gender-based violence cases in electronic medical records. Next,  these findings are complemented with a qualitative analysis of the patterns identified by this approach.

\subsection{Qualitative Evaluation: Models and Domains} 

\begin{figure} [h]
    \centering
    \includegraphics[width=1\linewidth]{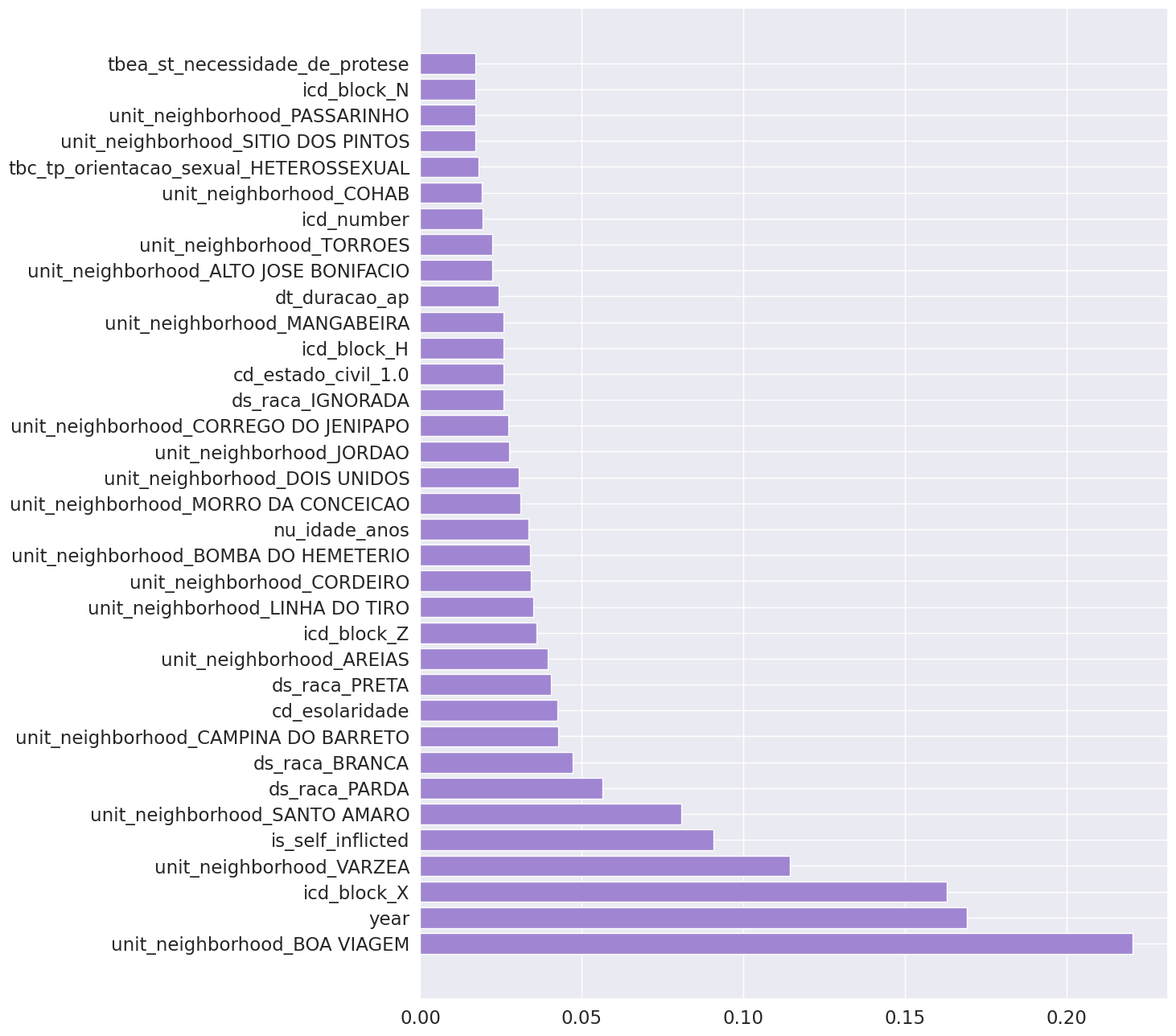}
    \caption{Most relevant features of the Demographic Model}
    \label{fig:DM}
\end{figure}

\paragraph{\textbf{Demographic Model}}
The list of the 35 most relevant features for the demographic model consists of the fill-in options associated with the parameterized fields, as shown in Figure \ref{fig:DM}. These features provide limited insight, as –– at this stage –– it is not possible to determine whether they contributed to the classification of violence or non-violence cases.

The feature of greatest relevance is the neighborhood in which a health unit is located. In fact, 19 of the 35 selected features are related to the location of healthcare facilities. The location of a health unit does not have a clear explanatory link to the occurrence of violence, particularly since patients may seek care in different facilities, making geographic information a weak and potentially misleading indicator.

Race also appears as a highly relevant feature. Of the six possible values, four — \textit{branca} (‘white’), \textit{parda} (‘brown’), \textit{preta}(‘black’), and \textit{ignorada} (‘ignored’) — were selected, with three among the top ten. This pattern can be influenced by inconsistent field completion, which can amplify the weight of filled values. As a result, the model emphasizes individual characteristics rather than contextual information, reinforcing stereotypes this work aims to avoid.

Despite these limitations, some relevant features are associated with the type of care sought, notably ICD blocks X, Z, H, and N. These correspond to external causes of morbidity and mortality, factors influencing health status and contact with health services, diseases of the eye and adnexa, and diseases of the genitourinary system. Although not predominant, the presence of blocks N and Z is consistent with the patterns identified by the semantic model, as block N may relate to sexual violence and block Z often reflects scenarios requiring follow-up care, such as prenatal monitoring, that are aligned with the findings in the semantic setup.

Thus, this qualitative analysis reinforces the quantitative findings. Parameterized data leads to a model focused primarily on individual attributes, with limited attention to the clinical context. Meaningful interpretation of care-related features was only possible through a comparison with the semantic model, further supporting the use of FrameNet-based semantic analysis for open-text fields. The next section examines the most relevant features of the semantic model that had the best performance in the quantitative analysis.

\paragraph{\textbf{Semantic Model}}

\begin{figure} [h]
    \centering
    \includegraphics[width=1\linewidth]{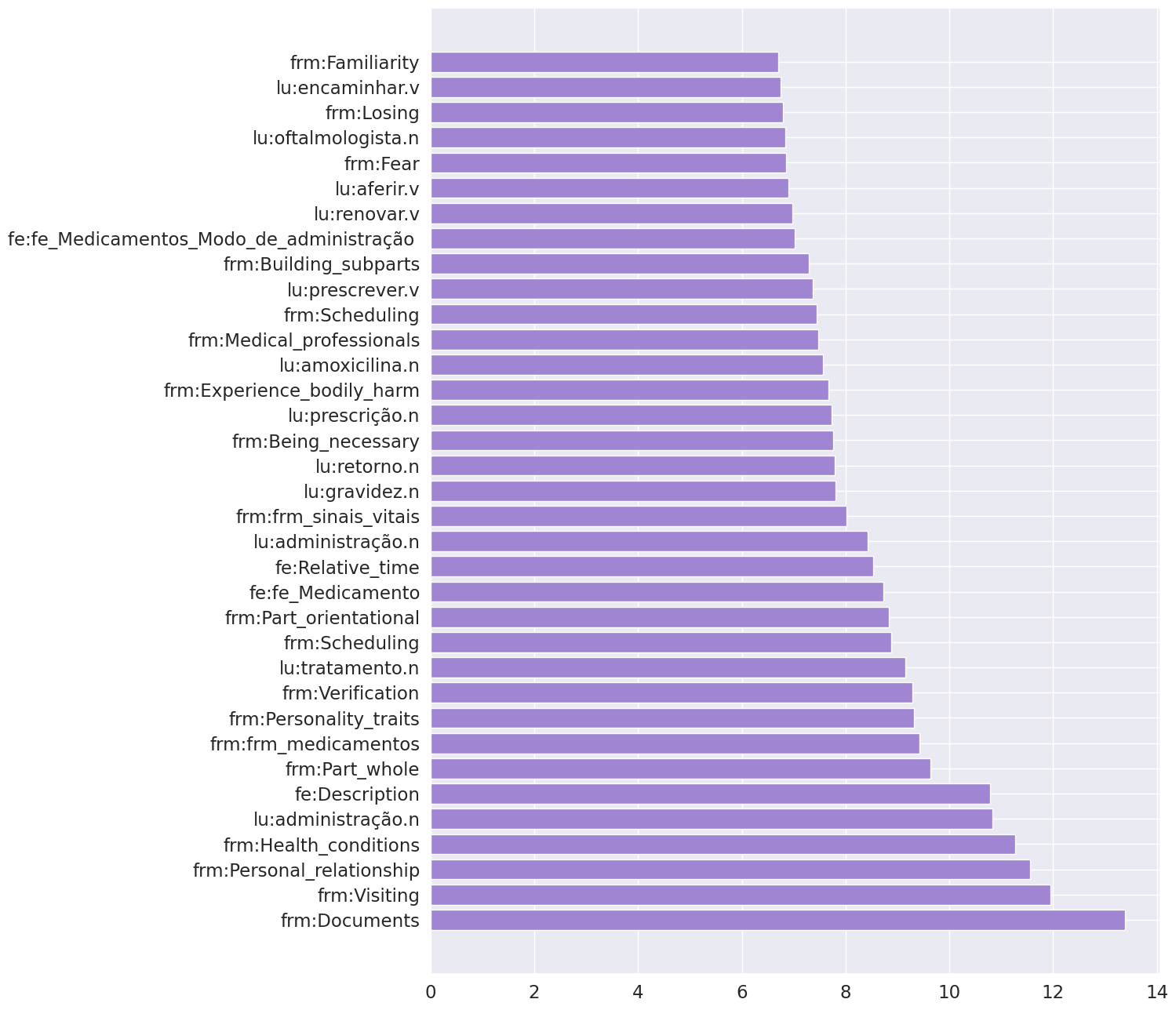}
    \caption{Most Relevant Features of the Semantic Model}
    \label{fig:SM}
\end{figure}

 Figure \ref{fig:SM} shows the features that most influenced the semantic model, including frames, frame elements, co-occurrences between frame elements, and lexical units. At this stage, the analysis is exploratory and the discussion focuses on recurring patterns rather than on a detailed interpretation.
 
Relevant features are not restricted to the Healthcare and Violence domains. Among the most influential features are general vocabulary frames, most notably \texttt{Personal\_relationships}. This result is unsurprising, given that many cases of gender-based violence involve individuals who are related in some way, which may reflect indirect references to aggressors in the records. Another generic frame that appears prominently is \texttt{Fear}. Although it is not part of the Violence domain, its relevance suggests that emotional states expressed in the text may contribute to identifying situations associated with violence.

Healthcare-related frames are more frequent than violence-related ones. Only the \texttt{Experience\_bodily\_harm} frame appears as a relevant frame from the Violence domain. However, the presence of this frame supports the idea that violence-related patterns can be inferred from clinical information. This is particularly relevant for primary care settings, where early identification is critical.

Within the Healthcare domain, \texttt{Health\_Conditions} ranks among the most relevant frames, which is expected given the nature of the data. Frames related to \texttt{Medicines} also appear repeatedly, including more than one of their frame elements. In addition, several lexical units associated with the \texttt{Health\_Intervention} frame appear as relevant for the model’s decisions. Together, these results indicate that routine clinical actions and treatment-related information play an important role in the identification process.

The semantic features identified in the model provide a starting point for understanding how patterns related to gender-based violence can appear in clinical narratives. Rather than examining all relevant patterns, the focus was on the most frequently evoked frames –– and lexical units –– annotated by LOME in the confirmed cases of GBV that are part of the Healthcare and Violence domains.

Within the Healthcare domain, \texttt{Health\_conditions} is by far the most frequently evoked frame. This is expected given the nature of clinical records and it was also evident in the model's features' analysis. However, only a small number of its associated lexical units appeared among the most frequent terms, suggesting that not all health conditions carry the same analytical weight. One term that stands out is \textit{gestante.n} (‘pregnant’), while \textit{gravidez.n} ('pregnancy') was also a relevant feature for the semantic model. The prominence of pregnancy-related terms raises an interpretive challenge: it is unclear whether this reflects increased vulnerability, patterns specific to the dataset, or a potential bias toward the specific gender healthcare event in focus.

The \texttt{Health\_service} frame also plays an important role. Terms such as \textit{encaminhar.v} (‘referral’) and \textit{acompanhamento.n} (‘follow-up’) occur frequently and often refer to specialized care or ongoing treatment. When considered together with other clinical elements, these references may indirectly signal previous incidents. Similarly, frequent mentions of medical examinations and medications suggest that routine clinical procedures may carry contextual clues. Enriching these elements through semantic relations, such as ternary qualia relations, may allow deeper inferences about the underlying conditions and possible links to violence.

A closer look at the \texttt{Health\_conditions} frame reveals two notable tendencies. First, pregnancy-related terms appeared with high frequency, as it was already pointed out.  Second, mental health conditions –– including depression, anxiety, and bipolar disorder –– were highly represented. These patterns may reflect the psychological consequences of abuse, although they may also be influenced by broader gendered healthcare-seeking behaviors. In either case, they warrant further investigation.

In contrast, frames from the Violence domain appear less frequently, likely because, in comparison to health issues, explicit references to violence are less frequent in clinical records. Among them, the \texttt{Experience\_bodily\_harm} frame stands out and also contributed to the performance of the model. However, many of its associated terms, such as fall or trauma, are not inherently indicative of violence. More direct signals emerge in references to self-inflicted harm, including self-mutilation and suicide. Although these cases were not analyzed separately, their frequency suggests that self-directed violence deserves closer attention in future work.

Sexual violence-related patterns are particularly salient. Terms associated with sexual acts, abuse, and related examinations appear consistently, indicating that sexual violence may be a significant factor motivating healthcare visits. References to sexually transmitted infections further reinforce this interpretation. These patterns suggest that even when violence is not explicitly documented, its consequences may be traceable through clinical descriptions.

Therefore, this qualitative analysis also shows that FrameNet-based semantic annotation makes it possible to uncover patterns that would likely remain invisible in structured data alone. Although the findings remain exploratory and require validation in collaboration with healthcare professionals, they support the broader hypothesis that semantic analysis of open-text medical records can contribute to the identification of gender-based violence in primary care settings.

\section{Conclusion and Future Work}

This study evaluated the use of FrameNet-based semantic annotation to identify Gender-Based Violence (GBV) cases and patterns in open-text fields of e-medical records. Our results show that models using semantic annotation of open-text fields outperform models relying solely on parameterized demographic data, achieving an F1 score 0.31 higher. The addition of parameterized fields to the semantic model provided minimal improvement, highlighting that open-text information carries richer and more relevant insights for detecting GBV. Qualitative analysis confirmed that relying only on parameterized data risks reinforcing stereotypes and provides limited information for pattern discovery, while semantic annotation enables the identification of meaningful patterns that can inform further investigation and policy intervention. 

In general, these findings support the hypothesis that FrameNet-based semantic analysis is a valuable tool for identifying potential GBV cases, including those underreported. By revealing patterns in both reported and unreported cases, this approach can assist in early-warning systems and public policies, contributing to improved protection and intervention strategies. Finally, the experimental setups and qualitative assessments presented here provide a baseline for future research on leveraging linguistic analysis for public health surveillance, which is already in development. Progress has already been made on the more systematic identification of GVB patterns \cite{dutra2025framesemanticpatternsidentifying}, and three parallel lines of research are currently being carried out to explore the identification of new patterns. Two of them focus directly on violence-related cases, specifically self-harm and violence against LGBTQ+ individuals. The third broadens the scope of the semantic model beyond violence, with the aim of identifying patterns related to women’s health and supporting early detection of potential cancer cases.

\section*{Ethics and Limitations}

The study presented in this paper was part of a broader project and approved by the Research Ethics Committee of the Federal University of Goiás (CAAE:64733922.3.0000.5083; Approval number: 6.126.995). 
The research involved highly sensitive information from violence notifications and electronic medical records that could increase the risk for victims of violence. Thus, the research team has extensively studied this issue and consulted data protection specialists before pursuing any implementation of the methodology. To protect the information, all team members signed confidentiality agreements, the data was anonymized - as described-, and access was restricted to anonymized samples only.
The methodology was developed to improve the use of health data in Brazil and address the underreporting of health-related events, using frame-based modeling, semantic parsing, and the identification of linguistic pattern in Brazilian Portuguese.  Although it can be adjusted and expanded to other languages, it has not been extensively tested  yet and may reflect biases specific to Brazilian Portuguese, which is a limitation.


\section*{Acknowledgments}

This work was supported by the Patrick J. McGovern Foundation’s acceleration program, the José Luiz Setúbal Foundation, and the Instituto Galo da Manhã. We also express our gratitude to our partners from the Recife Municipal Health Department — Luciana Caroline, Marcella Abath, Natalia Barros, and Yana Lopes — for their valuable collaboration and continuous support throughout this project. Tiago Torrent is a grantee of the Brazilian National Council for Scientific and Technological Development CNPq – grant 311241/2025-5).

\bibliography{custom}




\end{document}